\documentclass[runningheads]{llncs}
\usepackage{graphicx}
\usepackage{amsmath,amssymb} 
\usepackage{color}
\usepackage[width=122mm,left=12mm,paperwidth=146mm,height=193mm,top=12mm,paperheight=217mm]{geometry}
\begin{document}
\pagestyle{headings}
\mainmatter
\def\ECCV16SubNumber{1207}  

\institute{Mitsubishi Electric Research Labs}

\title{Global-Local Face Upsampling Network} 

\titlerunning{Global-Local Face Upsampling Network}

\authorrunning{Tuzel, Taguchi, and Hershey}

\author{Oncel Tuzel\ \ \ \ \ \ \  Yuichi Taguchi\ \ \ \ \ \ \  John R. Hershey}

\institute{Mitsubishi Electric Research Labs (MERL), Cambridge, MA, USA}

\def\etal{et al.}

\maketitle

\begin{abstract}
Face hallucination, which is the task of generating a high-resolution face image from a low-resolution input image, is a well-studied problem that is useful in widespread application areas. Face hallucination is particularly challenging when the input face resolution is very low (e.g., $10 \times 12$ pixels) and/or the image is captured in an uncontrolled setting with large pose and illumination variations. In this paper, we revisit the algorithm introduced in~\cite{Liu07} and present a deep interpretation of this framework that achieves state-of-the-art under such challenging scenarios. In our deep network architecture the global and local constraints that define a face can be efficiently modeled and learned end-to-end using training data. Conceptually our network design can be partitioned into two sub-networks: the first one implements the holistic face reconstruction according to global constraints, and the second one enhances face-specific details and enforces local patch statistics. We optimize the deep network using a new loss function for super-resolution that combines reconstruction error with a learned face quality measure in adversarial setting, producing improved visual results. We conduct extensive experiments in both controlled and uncontrolled setups and show that our algorithm improves the state of the art both numerically and visually.


\end{abstract}

\section{Introduction} \label{sec:intro}

Face has been one of the main targets for image enhancement tasks. In particular, upsampling a low-resolution face image to a high-resolution one has been an important problem, called by its own name of face hallucination~\cite{Baker00FG}.

The problem is formulated as follows. Given a low-resolution $N_L = n \times m$ face image $I_L$, our goal is to obtain a photorealistic high-resolution $N_H = (d  n) \times (d  m)$ face image $I_H$ whose down-sampled version is equal to $I_L$, where $d$ is the upsampling factor. The relation can be written as
\begin{equation}\label{eq:lr_hr}
 \mathbf{x}_L = \mathbf{K} \; \mathbf{x}_H ,
\end{equation}
where $\mathbf{x}_L$ and $\mathbf{x}_H$ are the low and high-resolution images stacked into column vectors and $\mathbf{K}$ is an $N_L \times N_H$ sparse matrix implementing low-pass filtering and down-sampling. To invert this largely ($d^2$-times) under-determined linear system and recover the high-resolution image, additional constraints are needed.

We approximate the solution of this linear inverse problem using a deep neural network where facial constraints are explicitly modeled and learned using training data.
Our considerations for the proposed face upsampling network are inspired by the face hallucination work of Liu~\etal~\cite{Liu07}. Similar to~\cite{Liu07}, we utilize the following three constraints to regularize the under-determined problem. (1) Global constraint: The reconstructed high-resolution face image should satisfy holistic constraints such as shape, pose, and symmetry, and should include detailed characteristic facial features such as eyes and nose. (2) Local constraint: Statistics of the reconstructed local image regions should match that of high-resolution face image patches (e.g., smooth regions with sharp boundaries), and should include face-specific details. (3) Data constraint: The reconstruction should be consistent with the observed low-resolution image and satisfy Eq.~\eqref{eq:lr_hr}.

Liu~\etal~\cite{Liu07} used a two-step approach according to these constraints. First a global face reconstruction is acquired using an eigenface model, which is a linear projection operation. In the second step details of the reconstructed face are enhanced by non-parametric patch transfer from a training set where consistency across neighboring patches is enforced through a Markov random field. This method produces high-quality results when the face images are near frontal, well aligned, and lighting conditions are controlled. However, when these assumptions are violated, the simple linear eigenface model fails to produce satisfactory global reconstruction. In addition, the patch transfer does not scale well with large datasets due to the nearest-neighbor (NN) search.

In this paper, we present a deep network architecture that resembles Liu~\etal's framework~\cite{Liu07} but solves the aforementioned problems for accurate and efficient face hallucination. Our network consists of the two sub-networks: the first one implements holistic face reconstruction according to the global constraints, and the second one enhances face-specific details and enforces local patch statistics. However, they are learned jointly using a large amount of training data, providing an optimized structure for upsampling. Moreover, the feed-forward operation provides computational efficiency in the test time. In extensive experiments using two benchmark datasets captured under controlled and uncontrolled setups, we show that our algorithm outperforms the state-of-the-art algorithms.

\subsection{Contributions}

Our main contributions can be summarized as follows: (1) We present a deep interpretation of the global-local face hallucination framework~\cite{Liu07}. (2) We design a deep network architecture that replaces the original two-step approach with an end-to-end learning and feed-forward operation, improving both the accuracy and speed. (3) We learn the deep network by minimizing a combination of reconstruction error and a learned face quality loss in adversarial setting, which produces high resolution images with improved visual quality. (4) We conduct extensive comparisons with the state-of-the-art algorithms and demonstrate that our algorithm outperforms them both qualitatively and quantitatively.

\subsection{Related Work}

Face hallucination is the single-image super-resolution (SR) problem specific to face images. Single-image SR algorithms developed for generic images share the same formulation in (\ref{eq:lr_hr}). To invert the under-determined system, local constraints are enforced as priors based on image statistics~\cite{fattal2007image,sun2008image,Kim10PAMI} and exemplar patches~\cite{freeman2002example,glasner2009super,yang2010image}. Global constraints are typically not available for the generic SR problem, which limits the plausible upsampling factor. Yang~\etal's recent study~\cite{Yang14ECCV} showed that $4\times$ upsampling results in the lower bound of the human perceptual scores.

Liu~\etal~\cite{Liu07} used a global constraint for face hallucination based on eigenfaces~\cite{turk1991eigenfaces}, and proposed a two-step approach where the initial global reconstruction is improved by local non-parametric patch transfer~\cite{freeman2002example}. As described above, the simple eigenface model has a difficulty when the datasets include large pose and illumination variations, and their local refinement process is computationally expensive due to the NN patch search.
Ma~\etal~\cite{Ma10PR} assumed that training and test images are precisely aligned and searched the NN patches of a target pixel in the test image only at the specific pixel location in the training images. Using the location-specific patches provides global constraints implicitly, as long as the images are well-aligned.
Yang~\etal~\cite{Yang13CVPR} partitioned a face image into three groups of facial components, contours, and smooth regions based on a facial landmark detection~\cite{Zhu12CVPR}. They used the NN search for each of the facial components with the training images, while for contours and smooth regions edge-based statistics and NN patch search were used. The result was generated by integrating gradient maps from the three groups and imposing them on the high-resolution image. Their method relies on the facial landmark detection and thus the result degrades for low-resolution input images where the landmark localization is typically inaccurate.

In the past several years, the success of deep learning methods has revolutionized the computer vision field from image classification~\cite{krizhevsky2012imagenet,simonyan2014very} and object detection~\cite{girshick2014rich} to face recognition~\cite{taigman2014deepface}, segmentation~\cite{farabet2013learning}, and video event detection~\cite{simonyan2014two}. These methods have also been replacing highly optimized hand-designed algorithms in low-level vision tasks such as image denoising~\cite{burger2012image,agostinelli2013adaptive,jain2009natural}, image enhancement~\cite{xie2012image,xu2014deep}, and SR~\cite{Dong15PAMI,wang2015self,Zhou15AAAI}.
Dong~\etal~\cite{Dong15PAMI} proposed super-resolution convolutional neural network (SRCNN) for generic SR. They interpreted it as a deep network version of the conventional sparse coding methods~\cite{yang2010image}. SRCNN provides the state-of-the-art performance for generic SR, but not for face-specific SR as we compare in experiments. More recently, Wang~\etal~\cite{wang2015self} proposed an improved deep model for generic SR that also takes into account self similarities.
Zhou~\etal~\cite{Zhou15AAAI} presented bi-channel convolutional neural network (BCCNN) for face-specific SR. They used a convolutional neural network architecture whose output was blended with the bicubic upsampled image using a weighting factor which is also predicted from the network.
The last layer of this network linearly combines high-resolution basis images, which corresponds to a global face reconstruction and smooths out the person-specific details.

Basic building blocks of our algorithm are well known neural network architectures such as encoder~\cite{Hinton06,Vincent08,ngiam2011multimodal}, convolutional~\cite{lecun1998gradient}, and deconvolutional~\cite{zeiler2010deconvolutional,Long15} neural nets. Our architectural design enables effective learning of global and local constraints that are important for face upsampling task using these well-known building blocks.

Recently, generative adversarial networks (GANs)~\cite{Goodfellow14} have been proposed as an alternative to learn deep generative models. In GAN framework, a generative network learns to generate samples from a given data distribution, while simultaneously a discriminative network learns to identify the samples that are generated from this network. Since then, GANs have been successfully used for image~\cite{Goodfellow14,Denton15}, scene~\cite{Radford15}, and sequence synthesis~\cite{Lotter15} tasks. In this paper, we use GAN framework to learn a discriminative network which evaluates face quality, while at the same time optimizing the face super-resolution network according to the learned quality measure.

\section{Method Overview}

\begin{figure}[tb]
  \centering
  \includegraphics[width=0.75\textwidth, trim={0.4in 0.65in 3.1in 0.1in},clip]{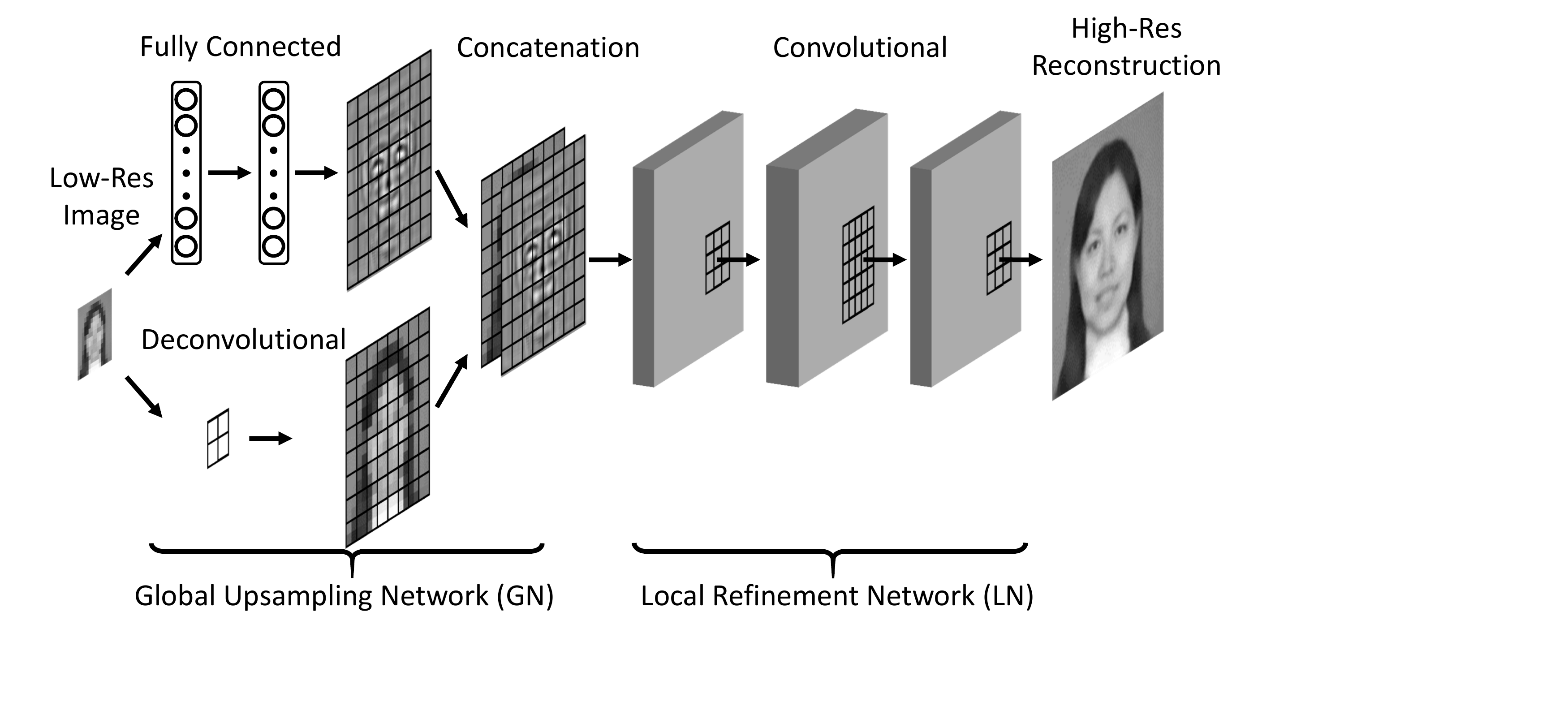}
  \vspace{-3mm}
  \caption{Our global-local face upsampling network (GLN).}
\label{fig:Overview}
\end{figure}

Figure~\ref{fig:Overview} shows an overview of our global-local face upsampling network (GLN). Our network consists of two sub-networks, referred to as Global Upsampling Network (GN) and Local Enhancement Network (LN), which model the global and local constraints for face hallucination. The operations performed by the two sub-networks are conceptually similar to Liu~\etal's global reconstruction followed by local enhancement~\cite{Liu07}.
However, by jointly modeling and learning global and local constraints using a deep architecture, our method provides higher accuracy while being more efficient in the test time due to the feed-forward processing, as detailed below.

Holistic face reconstruction according to global constraints is achieved using GN, which is a two stream neural network running in parallel. The first stream implements a simple interpolation-based upsampling of the low-resolution face using a deconvolutional network, producing a smooth image without details. The second stream produces the high frequency characteristic facial details, such as eyes and nose, using a fully connected neural network. Hidden layers of this encoder network~\cite{Hinton06,Vincent08} build a global representation of high-resolution face images that can be inferred from the low-resolution input. Compared to the linear eigenface model of~\cite{Liu07}, the multi-layer nonlinear embedding and reconstruction used in our framework enables  more effective encoding of characteristic facial features, in addition to variations such as alignment, face pose, and illumination. We concatenate the two streams generated by GN to be processed by LN.

The local constraints are modeled in LN using a fully convolutional neural network, which implements a shift-invariant nonlinear filter. This network enhances the face-specific local details by fusing the smooth and detail layers produced by the GN. Even though the convolutional filters are relatively small ($5 \times 5$ or $7 \times 7$), by stacking many filters, the receptive field of the network becomes quite large ($43$ pixels in 8 layer network). The large receptive field enables resolving the ambiguity (e.g., eye region vs. mouth region) and the deep architecture has enough capacity to apply necessary filtering operation to a given region. Compared to the non-parametric detail transfer of~\cite{Liu07}, our structure is very efficient and produces higher quality enhancement.

Although we do not explicitly model the data constraint within the network, by training the network using a large amount of training data, the network learns to produce high-resolution face images that are consistent with the low-resolution images according to Eq.~\eqref{eq:lr_hr}. One could enforce the data constraint by using the back-projection (BP) algorithm~\cite{Irani91} in a postprocessing step, which is a common approach used in many upsampling schemes. However, in our experiments, we found that the results directly obtained from our network and those obtained after the BP postprocessing were indistinguishable both qualitatively and quantitatively. Thus we did not use such a postprocessing.

\section{Global-Local Upsampling Network (GLN)} \label{sec:Upsampling}

This section presents the details of our deep network architecture that is used to upsample the low-resolution face images.
Our network structure is designed for very low-resolution input face images. We consider two upsampling factors: (1) $4 \times$ upsampling where  $32 \times 32$ input face image is mapped to $128 \times 128$ resolution; (2) $8 \times$ upsampling where  $16 \times 16$ input face image is mapped to $128 \times 128$ resolution. We have two different network configurations for the $4 \times$ and $8 \times$ cases. We assume that the low-resolution face images are roughly aligned.

\subsection{Global Upsampling Network (GN)}

The structure of the GN is summarized in Table~\ref{tab:GN}. GN is a two stream network running in parallel. The image upsampling stream maps the input face image to a high-resolution face image using linear interpolation. In our network, we implemented the image upsampling stream using a deconvolution layer~\cite{zeiler2010deconvolutional,Long15}. We initialize the interpolation weights using bilinear matrix but allow the weights to change during training.

The global detail generation stream is implemented as a fully connected encoder network with 3 hidden layers. We use rectified linear unit (ReLU) after every linear map except for the last layer which generates the $128 \times 128$-dimensional upsampled global detail. In our encoder network the code layer is $256$-dimensional, both for $4 \times$ and $8 \times$ upsampling networks. This is mainly dictated by the limited amount of training data where larger latent spaces for the global detail tend to overfit. Finally we concatenate the outputs of the image upsampling stream and the global detail generation stream, and form a $2 \times 128 \times 128$ tensor to be processed by the LN.

Figure~\ref{fig:Network} shows a typical output of the $8 \times$ GN. Even though we allow the image upsampling stream's weights to change during training, the weights tend not to change much and the network implements a smooth upsampling (Figure~\ref{fig:Network}(a)). The output of the global detail generation stream (Figure~\ref{fig:Network}(b)) encodes high frequency details and more difficult to interpret. The pattern is more visible around the characteristic facial features such as eyes, nose and mouth.

\begin{table}[tb]
	\centering
	\caption{Global upsampling network architecture. First and second columns are the image upsampling and global detail generation streams for $4 \times$ image upsampling. Third and fourth columns are the corresponding streams for the $8 \times$ image upsampling.}
	\label{tab:GN}
	\vspace{-4mm}
		\begin{tabular}[t]{|c|c|c|c|}
		\hline
		\multicolumn{2}{|c} {$4 \times$ GN } & \multicolumn{2}{|c|} {$8 \times$ GN } \\ \hline \hline
		deconv4 & fc-512 & deconv8 & fc-256 \\ 
						& fc-256 & 			  & fc-256 \\ 
						& fc-512 & 			  & fc-256 \\ 
						& fc-($128 \times 128$) & 			  & fc-($128 \times 128$) \\ \hline
		\multicolumn{2}{|c} {concatenation } & \multicolumn{2}{|c|} {concatenation } \\ \hline			
		\end{tabular}
\end{table}

\begin{table}[tb]
	\centering
	\caption{Local refinement network architecture. The first number after conv. indicates the kernel size whereas the second number is the number of filters. We analyzed three convolution architectures with 4, 6, and 8 layers. The same architectures are used for both $4 \times$ and $8 \times$ upsampling tasks.}
    \vspace{-4mm}
	\label{tab:LN}
		\begin{tabular}[t]{|c|c|c|}
		\hline
		4 Layer LN (LN4) & 6 Layer LN (LN6) & 8 Layer LN (LN8)\\ \hline \hline
		conv5-16 & conv5-16 & conv5-16 \\
		conv7-64 & conv7-32 & conv7-32 \\
		conv5-16 & conv7-64 & conv7-64\\
						 & conv7-32 & conv7-64\\
						 & conv5-16 & conv7-64\\						
						 & 				  & conv7-32\\												
						 & 				  & conv5-16\\ \hline
		conv5-1  & conv5-1  & conv5-1\\ \hline								
		\end{tabular}
\end{table}

\subsection{Local Refinement Network (LN)}

\begin{figure}[!tb]
  \centering
	\small
\setlength{\tabcolsep}{0.4mm}
	\begin{tabular}{ccc}	
  \includegraphics[width=.2\textwidth]{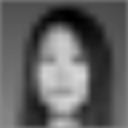} &
	\includegraphics[width=.2\textwidth]{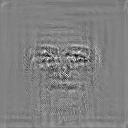} &
	\includegraphics[width=.2\textwidth]{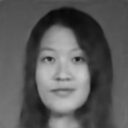} \\
	(a) & (b) & (c)
	\end{tabular}
    \vspace{-2mm}
  \caption{Example output of the (a) image upsampling and (b) global detail generation streams of the $8 \times$ GN. (c) Final upsampling result after LN.}
\label{fig:Network}
\end{figure}

The structure of the LN is summarized in Table~\ref{tab:LN}. These structures are identical for $4 \times$ and $8 \times$ upsampling tasks. We analyzed three fully convolutional neural network architectures with different numbers of layers (LN4, LN6, and LN8). Before each convolution operation the image is padded with the ceiling of the half filter size so that the output image dimension is same as the input dimension. After every convolutional layer we apply ReLU except the last layer which constructs the final upsampled image. We do not perform pooling and the stride is 1, therefore this network learns a very large shift-invariant nonlinear filter. As shown in Figure~\ref{fig:Network}(c), the LN enhances the face specific local details by fusing the smooth and detail layers produced by the GN (see eyes and nose). In addition, the reconstructed image's local statistics match that of high-resolution face image patch statistics (e.g., smooth cheek region and sharp face boundaries).

\subsection{Training} \label{sec:Training}

We conducted two stage training procedure. In the first stage, we train the network by minimizing a reconstruction error criteria. During the (optional) second stage, we fine-tune the network by minimizing a weighted combination of reconstruction error and a learned face quality loss function.

{\bf Training for reconstruction:}
We minimize the mean-squared loss between the ground truth high-resolution images and the reconstructed images to learn the network parameters. Let $\{(\mathbf{x}_L^i , \ \mathbf{x}_H^i)\}_{i=1,\ldots,n}$ be the set of $n$ low-resolution and high-resolution training image pairs. The loss is given by
\begin{equation}
 L_{MS} = \frac{1}{n} \sum_{i=1}^n \|G(\mathbf{x}_L^i) -  \mathbf{x}_H^i\|^2,
\end{equation}
where $G(.)$ is the GLN function. The network is trained using stochastic gradient descent (SGD) on mini-batches of $5$ images, with fixed learning rate schedule $10^{-8}$, and momentum $0.9$.

{\bf Adversarial fine-tuning:}
Mean-squared loss function prefers blurry reconstructions at the regions with high ambiguity, such as edges. Here we complement this loss function with a learned loss function, which is tuned to measure the quality of reconstruction by discriminating reconstructed images from the true high resolution images. We use a variant of the generative adversarial network framework proposed by Goodfellow~\etal~\cite{Goodfellow14} to learn the discriminative loss function in conjunction with the GLN parameters.

In our framework, the discriminative network $D(.)$ detects the images reconstructed by the GLN. It is trained to maximize the output probability when the input is reconstructed by the GLN and minimize the output probability when the input is a true high resolution face image using the loss function:
\begin{equation} \label{eq:discriminator}
 L_{D} = -\frac{1}{2n} \sum_{i=1}^n \big( \log (1 - D(\mathbf{x}_H^i)) + \log (D(G(\mathbf{x}_L^i))) \big).
\end{equation}

The GLN is trained both to minimize the reconstruction error $L_{MS}$, and to confuse the discriminative network by minimizing the output probability of the discriminative network on the input reconstructed images. This is achieved by using the combined mean-squared and adversarial loss function:
\begin{equation} \label{eq:generator}
 L_{G} =  L_{MS} - \lambda \frac{1}{n} \sum_{i=1}^n \log (1 - D(G(\mathbf{x}_L^i))),
\end{equation}
where $\lambda$ is a weighting factor between the two loss terms.

Similar to~\cite{Goodfellow14}, we alternate between minimizing $L_D$ with respect to parameters of the discriminative network, while keeping the GLN parameters fixed, and minimizing $L_G$ with respect to parameters of the GLN, while keeping discriminative network parameters fixed. We used $10$ SGD iterations for discriminative network and $50$ SGD iterations for the GLN during alternations. We switched 10000 times between optimizing discriminative network and the GLN.

The discriminative network was implemented as a convolutional neural network with four layers: (1) conv5-16, ReLU, MaxPool 2x2, (2)  conv5-16, ReLU,  MaxPool 2x2, (3) fc-50, ReLU, (4) fc-2. We started adversarial fine-tuning using the network trained for reconstruction only. The weighting factor $\lambda$ was set such that the initial adversarial loss was equal to $1/10$ of the mean-squared loss. We used open source CAFFE library~\cite{jia2014caffe} to implement the networks.

\section{Experiments} \label{sec:Exp}

We designed two sets of experiments under controlled and uncontrolled (in the wild) settings. The controlled setting was conducted using Face Recognition Grand Challenge (FRGC) dataset~\cite{phillips2005overview}, where frontal face images were taken in a studio setting under two lighting conditions with only two facial expressions (smiling and neutral). We used a total of $22,149$ images, where $20,000$ images were used for training and $2,149$ for testing. We used a variant of the supervised descent face alignment algorithm~\cite{xiong2013supervised} to detect facial landmarks. We then applied similarity transformation to the input face images (translation, rotation and uniform scaling) to approximately align detected eye and mouth center locations to a set of fixed points.

The uncontrolled setting was conducted using an aligned version of the Labeled Faces in the Wild dataset~\cite{huang2007labeled}, called Labeled Faces in the Wild-a (LFW-a)~\cite{wolf2011effective}. This dataset is intended for studying unconstrained face recognition problem and includes face images with various illumination, poses, and expressions. It contains $13,233$ face images from $1,680$ people, where we used $12,000$ for training and $1,233$ for testing.
In both settings we kept the identities of the people in the training and testing sets disjoint. Note that the alignment is quite noisy for both settings, particularly for the LFW-a dataset.

We evaluated two upsampling factors of $4\times$ and $8\times$. Faces occupied approximately $20 \times 24$ pixel area in the  $4\times$ upsampling case and $10 \times 12$ pixel area in the $8\times$ upsampling case. The low-resolution images were generated using the procedure described in~\cite{Yang14ECCV}, filtering the high-resolution images with a Gaussian blur kernel $\sigma$ followed by down-sampling. We used $\sigma = 1.2$ for $4\times$ down-sampling as suggested, and $\sigma = 2.4$ for $8\times$ down-sampling.

\subsection{Comparisons}

We compared our method with the state-of-the-art generic and face-specific SR algorithms. As the generic SR algorithms, we used Kim and Kwon's algorithm (KK)~\cite{Kim10PAMI} and the SRCNN algorithm~\cite{Dong15PAMI}, which are among the top performers in the evaluations reported in~\cite{Yang14ECCV,Dong15PAMI}. We used the implementations and pre-trained models (the 9-5-5 network for SRCNN) available on the authors' websites\footnote{We also retrained the SRCNN using our datasets. These models had almost identical PSNR ($0.02dB$ better for both $4 \times$ and $8 \times$ upsampling) to the pre-trained models.}. Since their algorithms were trained only up to $4\times$ upsampling factor, we performed $4\times$ upsampling followed by $2\times$ upsampling to generate $8\times$ upsampling results. As the face-specific SR algorithms, we used (1) Liu~\etal's algorithm (LSF)~\cite{Liu07}; (2) Ma~\etal's (MZQ)~\cite{Ma10PR}; (3) Yang~\etal's (YLY)~\cite{Yang13CVPR}; and (4) Zhou~\etal's BCCNN~\cite{Zhou15AAAI}. We used Yang~\etal's implementations~\cite{Yang13CVPR} for LSF, MZQ, and YLY, and our own implementation for BCCNN. 
All the face-specific SR algorithms were trained using the same sets of training images as ours.
Our results presented in this section, both qualitatively and quantitatively, were obtained by minimizing the reconstruction error only. Note that after adversarial fine-tuning, the visual quality improves while quantitative results change marginally.

\begin{figure*}[t]
\centering
\setlength{\tabcolsep}{0.4mm}
\scriptsize
\begin{tabular}[t]{ccccccccc}

\includegraphics[width=.105\textwidth]{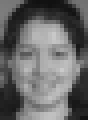} &
\includegraphics[width=.105\textwidth]{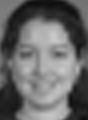} &
\includegraphics[width=.105\textwidth]{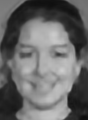} &
\includegraphics[width=.105\textwidth]{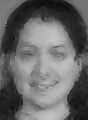} &
\includegraphics[width=.105\textwidth]{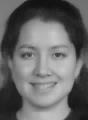} &
\includegraphics[width=.105\textwidth]{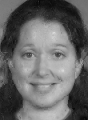} &
\includegraphics[width=.105\textwidth]{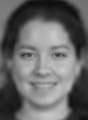} &
\includegraphics[width=.105\textwidth]{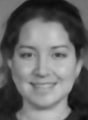} &
\includegraphics[width=.105\textwidth]{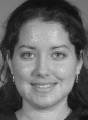} \\
\includegraphics[width=.105\textwidth]{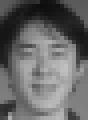} &
\includegraphics[width=.105\textwidth]{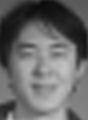} &
\includegraphics[width=.105\textwidth]{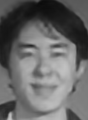} &
\includegraphics[width=.105\textwidth]{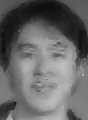} &
\includegraphics[width=.105\textwidth]{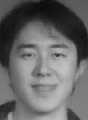} &
\includegraphics[width=.105\textwidth]{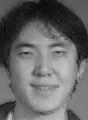} &
\includegraphics[width=.105\textwidth]{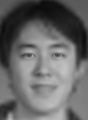} &
\includegraphics[width=.105\textwidth]{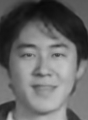} &
\includegraphics[width=.105\textwidth]{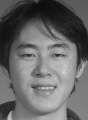} \\
\includegraphics[width=.105\textwidth]{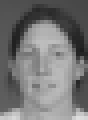} &
\includegraphics[width=.105\textwidth]{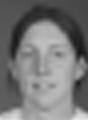} &
\includegraphics[width=.105\textwidth]{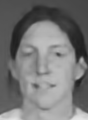} &
\includegraphics[width=.105\textwidth]{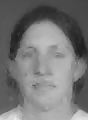} &
\includegraphics[width=.105\textwidth]{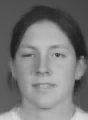} &
\includegraphics[width=.105\textwidth]{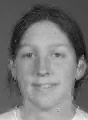} &
\includegraphics[width=.105\textwidth]{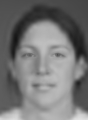} &
\includegraphics[width=.105\textwidth]{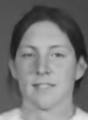} &
\includegraphics[width=.105\textwidth]{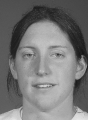} \\
\includegraphics[width=.105\textwidth]{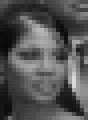} &
\includegraphics[width=.105\textwidth]{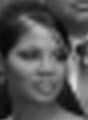} &
\includegraphics[width=.105\textwidth]{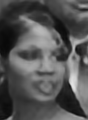} &
\includegraphics[width=.105\textwidth]{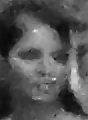} &
\includegraphics[width=.105\textwidth]{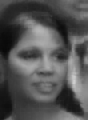} &
\includegraphics[width=.105\textwidth]{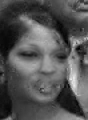} &
\includegraphics[width=.105\textwidth]{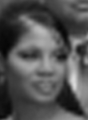} &
\includegraphics[width=.105\textwidth]{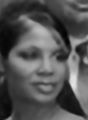} &
\includegraphics[width=.105\textwidth]{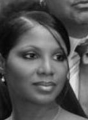} \\
\includegraphics[width=.105\textwidth]{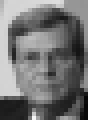} &
\includegraphics[width=.105\textwidth]{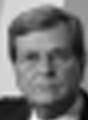} &
\includegraphics[width=.105\textwidth]{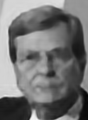} &
\includegraphics[width=.105\textwidth]{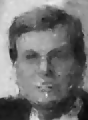} &
\includegraphics[width=.105\textwidth]{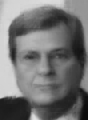} &
 &
\includegraphics[width=.105\textwidth]{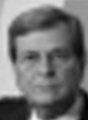} &
\includegraphics[width=.105\textwidth]{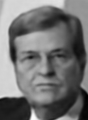} &
\includegraphics[width=.105\textwidth]{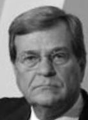} \\
\includegraphics[width=.105\textwidth]{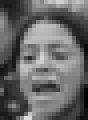} &
\includegraphics[width=.105\textwidth]{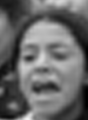} &
\includegraphics[width=.105\textwidth]{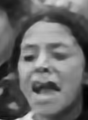} &
\includegraphics[width=.105\textwidth]{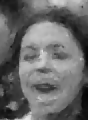} &
\includegraphics[width=.105\textwidth]{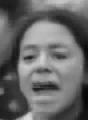} &
\includegraphics[width=.105\textwidth]{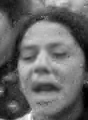} &
\includegraphics[width=.105\textwidth]{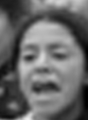} &
\includegraphics[width=.105\textwidth]{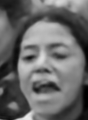} &
\includegraphics[width=.105\textwidth]{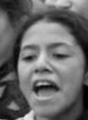} \\
NN &
Bicubic &
SRCNN &
LSF &
MZQ &
YLY &
BCCNN &
GLN &
GT \\

\end{tabular}
\caption{Qualitative comparisons of $4\times$ upsampling results for FRGC (top 3 rows) and LFW-a (bottom 3 rows) datasets. YLY does not produce results when the face landmarks cannot be detected in the low-resolution input. KK and SRCNN produced visually similar results, so the results obtained with KK are omitted.}
\label{fig:VisResults4x}
\end{figure*}

\begin{figure*}[tp]
\centering
\setlength{\tabcolsep}{0.4mm}
\scriptsize
\begin{tabular}[t]{ccccccccc}

\includegraphics[width=.105\textwidth]{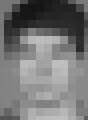} &
\includegraphics[width=.105\textwidth]{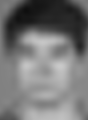} &
\includegraphics[width=.105\textwidth]{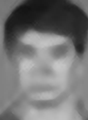} &
\includegraphics[width=.105\textwidth]{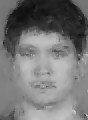} &
\includegraphics[width=.105\textwidth]{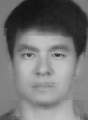} &
\includegraphics[width=.105\textwidth]{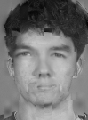} &
\includegraphics[width=.105\textwidth]{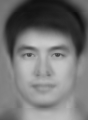} &
\includegraphics[width=.105\textwidth]{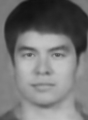} &
\includegraphics[width=.105\textwidth]{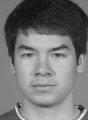} \\
\includegraphics[width=.105\textwidth]{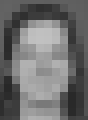} &
\includegraphics[width=.105\textwidth]{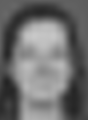} &
\includegraphics[width=.105\textwidth]{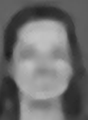} &
\includegraphics[width=.105\textwidth]{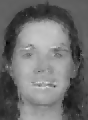} &
\includegraphics[width=.105\textwidth]{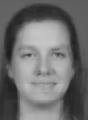} &
 &
\includegraphics[width=.105\textwidth]{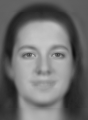} &
\includegraphics[width=.105\textwidth]{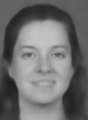} &
\includegraphics[width=.105\textwidth]{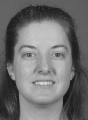} \\
\includegraphics[width=.105\textwidth]{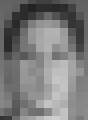} &
\includegraphics[width=.105\textwidth]{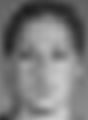} &
\includegraphics[width=.105\textwidth]{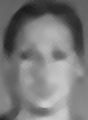} &
\includegraphics[width=.105\textwidth]{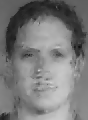} &
\includegraphics[width=.105\textwidth]{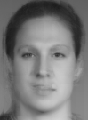} &
\includegraphics[width=.105\textwidth]{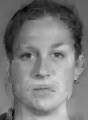} &
\includegraphics[width=.105\textwidth]{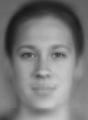} &
\includegraphics[width=.105\textwidth]{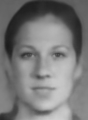} &
\includegraphics[width=.105\textwidth]{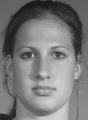} \\
\includegraphics[width=.105\textwidth]{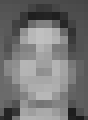} &
\includegraphics[width=.105\textwidth]{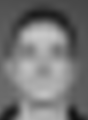} &
\includegraphics[width=.105\textwidth]{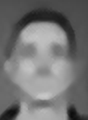} &
\includegraphics[width=.105\textwidth]{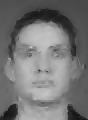} &
\includegraphics[width=.105\textwidth]{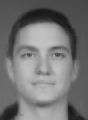} &
\includegraphics[width=.105\textwidth]{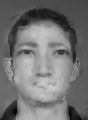} &
\includegraphics[width=.105\textwidth]{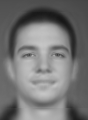} &
\includegraphics[width=.105\textwidth]{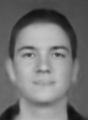} &
\includegraphics[width=.105\textwidth]{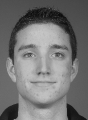} \\
\includegraphics[width=.105\textwidth]{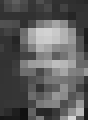} &
\includegraphics[width=.105\textwidth]{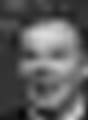} &
\includegraphics[width=.105\textwidth]{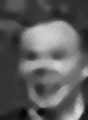} &
\includegraphics[width=.105\textwidth]{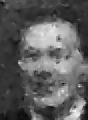} &
\includegraphics[width=.105\textwidth]{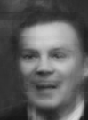} &
\includegraphics[width=.105\textwidth]{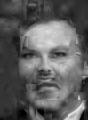} &
\includegraphics[width=.105\textwidth]{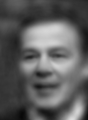} &
\includegraphics[width=.105\textwidth]{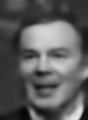} &
\includegraphics[width=.105\textwidth]{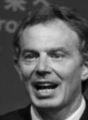} \\
\includegraphics[width=.105\textwidth]{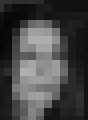} &
\includegraphics[width=.105\textwidth]{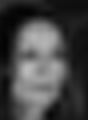} &
\includegraphics[width=.105\textwidth]{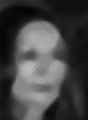} &
\includegraphics[width=.105\textwidth]{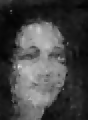} &
\includegraphics[width=.105\textwidth]{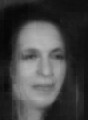} &
 &
\includegraphics[width=.105\textwidth]{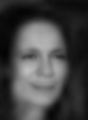} &
\includegraphics[width=.105\textwidth]{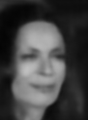} &
\includegraphics[width=.105\textwidth]{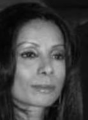} \\
\includegraphics[width=.105\textwidth]{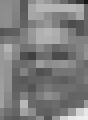} &
\includegraphics[width=.105\textwidth]{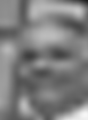} &
\includegraphics[width=.105\textwidth]{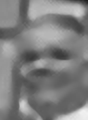} &
\includegraphics[width=.105\textwidth]{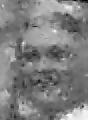} &
\includegraphics[width=.105\textwidth]{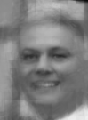} &
\includegraphics[width=.105\textwidth]{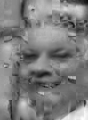} &
\includegraphics[width=.105\textwidth]{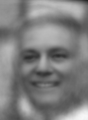} &
\includegraphics[width=.105\textwidth]{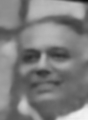} &
\includegraphics[width=.105\textwidth]{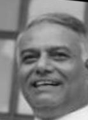} \\
\includegraphics[width=.105\textwidth]{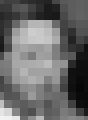} &
\includegraphics[width=.105\textwidth]{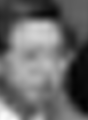} &
\includegraphics[width=.105\textwidth]{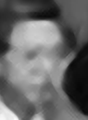} &
\includegraphics[width=.105\textwidth]{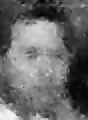} &
\includegraphics[width=.105\textwidth]{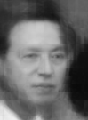} &
 &
\includegraphics[width=.105\textwidth]{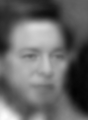} &
\includegraphics[width=.105\textwidth]{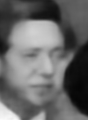} &
\includegraphics[width=.105\textwidth]{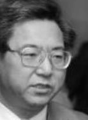} \\
NN &
Bicubic &
KK &
LSF &
MZQ &
YLY &
BCCNN &
GLN &
GT \\

\end{tabular}
\caption{Qualitative comparisons of $8\times$ upsampling results for FRGC (top 4 rows) and LFW-a (bottom 4 rows) datasets. YLY does not produce results when the face landmarks cannot be detected in the low-resolution input. KK and SRCNN produced visually similar results, so the results obtained with SRCNN are omitted.}
\label{fig:VisResults8x}
\end{figure*}

{\bf Qualitative Results:}
Figures~\ref{fig:VisResults4x} and \ref{fig:VisResults8x} respectively show $4\times$ and $8\times$ upsampling results. Note that our input resolution is low ($32\times32$ and $16\times16$ pixels), as it can be observed in the NN interpolation results. The bicubic interpolation results are blurry as expected. The generic SR algorithms (KK and SRCNN) sharpen the images by recovering some high-frequency components, but they do not reproduce face-specific features. The face-specific SR algorithms recover such facial features. In particular, we observed that MZQ and BCCNN produce visually pleasing results in our settings where the input resolution is low. However, since MZQ assumes precise alignments between the test and training images, the results degrade for the LFW-a dataset including larger alignment errors. Moreover, finding similar patches is hard for the case of $8\times$ upsampling, leading to inaccurate hallucination results for MZQ. BCCNN's network structure mainly performs global reconstruction by computing a weighted average of several high-resolution basis images and the bicubic upsampled image in the last layer of the network. This structure is similar to the GN-Only stream of our network, which is detailed in Section~\ref{sec:analysis}, and provides high resolution images with global details (e.g., symmetry and characteristic facial features) while suffering from loss of person-specific details. YLY, which relies on the facial landmark detection~\cite{Zhu12CVPR}, cannot produce results when the landmark detection fails; even if the detection is successful, the landmark localization accuracy is typically low for the low-resolution input, resulting in the facial features recovered at incorrect locations. We note that the resolution of our input images is lower than the images used in their original work. Our GLN produces globally consistent and locally sharp images, which are the closest to the ground truth (GT).

\begin{table}[t]
\centering
\small
\caption{Quantitative comparisons for FRGC $4\times$ and $8\times$ upsampling.}
\vspace{-4mm}
\label{tab:MetricsFRGC}
\begin{tabular}[t]{|c||c|c|c|c|c||c|c|c|c|c|}
\hline
Method & \multicolumn{5}{|c||}{FRGC $4\times$} & \multicolumn{5}{|c|}{FRGC $8\times$}\\
\cline{2-11}
 & PSNR & SSIM & IFC & WPSNR & NQM & PSNR & SSIM & IFC & WPSNR & NQM \\
\hline

NN                        & 25.39 & 0.694 & 1.25 & 35.17 & 6.94 & 21.95 & 0.466 & 0.39 & 28.79 & 3.49\\
\hline
Bicubic                   & 27.39 & 0.797 & 1.84 & 36.04 & 9.09 & 23.71 & 0.617 & 0.84 & 30.27 & 5.29\\
\hline
KK~\cite{Kim10PAMI}       & 28.06 & 0.825 & 2.10 & 37.34 & 9.72 & 24.11 & 0.631 & 0.89 & 30.85 & 5.74\\
\hline
SRCNN~\cite{Dong15PAMI}   & 28.19 & 0.829 & 2.10 & 37.47 & 9.91 & 24.18 & 0.641 & 0.91 & 30.89 & 5.84\\
\hline
LSF~\cite{Liu07}           & 25.64 & 0.723 & 1.28 & 33.56 & 7.42 & 23.21 & 0.614 & 0.76 & 30.35 & 4.90\\
\hline
MZQ~\cite{Ma10PR}         & 29.22 & 0.857 & 2.22 & 38.67 & 11.21 & 25.81 & 0.752 & 1.34 & 33.17 & 7.67\\
\hline
YLY~\cite{Yang13CVPR}     & 26.80 & 0.791 & 1.79 & 35.67 & 8.55 & 22.57 & 0.602 & 0.83 & 29.45 & 4.21\\
\hline
BCCNN~\cite{Zhou15AAAI}   & 28.67 & 0.842 & 2.13 & 37.98 & 10.63 & 25.70 & 0.749 & 1.34 & 32.90 & 7.60\\
\hline
GLN                       & \bf 30.34 & \bf 0.884 & \bf 2.66 & \bf 40.94 & \bf 12.37 & \bf 26.75 & \bf 0.787 & \bf 1.56 & \bf 34.37 & \bf 8.60\\
\hline

\end{tabular}
\end{table}

\begin{table}[t]
\centering
\small
\caption{Quantitative comparisons for LFW-a $4\times$ and $8\times$ upsampling.}
\vspace{-4mm}
\label{tab:MetricsLFWA}
\begin{tabular}[t]{|c||c|c|c|c|c||c|c|c|c|c|}
\hline
Method & \multicolumn{5}{|c||}{LFW-a $4\times$} & \multicolumn{5}{|c|}{LFW-a $8\times$}\\
\cline{2-11}
 & PSNR & SSIM & IFC & WPSNR & NQM & PSNR & SSIM & IFC & WPSNR & NQM \\
\hline

NN                        & 24.16 & 0.687 & 1.23 & 33.55 & 7.99 & 20.56 & 0.441 & 0.38 & 27.22 & 4.43\\
\hline
Bicubic                   & 26.62 & 0.796 & 1.84 & 34.61 & 10.72 & 22.16 & 0.575 & 0.77 & 28.38 & 6.13\\
\hline
KK~\cite{Kim10PAMI}       & 27.53 & 0.826 & 2.07 & 36.04 & 11.55 & 22.75 & 0.603 & 0.84 & 29.14 & 6.70\\
\hline
SRCNN~\cite{Dong15PAMI}   & 27.55 & 0.827 & 2.03 & 36.12 & 11.55 & 22.74 & 0.607 & 0.83 & 29.08 & 6.66\\
\hline
LSF~\cite{Liu07}           & 22.98 & 0.601 & 0.80 & 29.95 & 7.01 & 20.02 & 0.434 & 0.41 & 26.44 & 4.01\\
\hline
MZQ~\cite{Ma10PR}         & 26.36 & 0.784 & 1.61 & 34.10 & 10.69 & 22.64 & 0.621 & 0.83 & 29.11 & 6.89\\
\hline
YLY~\cite{Yang13CVPR}     & 25.52 & 0.750 & 1.54 & 33.62 & 9.51 & 20.80 & 0.500 & 0.59 & 27.30 & 4.82\\
\hline
BCCNN~\cite{Zhou15AAAI}   & 26.63 & 0.800 & 1.81 & 34.57 & 10.90 & 22.72 & 0.627 & 0.90 & 29.08 & 6.89\\
\hline
GLN                       & \bf 28.82 & \bf 0.863 & \bf 2.35 & \bf 37.80 & \bf 13.01 & \bf 24.07 & \bf 0.688 & \bf 1.12 & \bf 30.75 & \bf 8.19\\
\hline

\end{tabular}
\end{table}

{\bf Quantitative Results:}
In addition to the standard peak signal-to-noise ratio (PSNR) and structural similarity (SSIM), we computed the information fidelity criterion (IFC)~\cite{Sheikh05IFC}, weighted PSNR (WPSNR), and noise quality measure (NQM)~\cite{Damera00NQM} for quantitative evaluations as suggested in~\cite{Yang14ECCV}\footnote{The multi-scale SSIM (MSSSIM)~\cite{Wang03MSSSIM} with the default number of scales was not applicable due to the small output resolution ($128\times128$ pixels).}.
Tables~\ref{tab:MetricsFRGC} and \ref{tab:MetricsLFWA} show the results for different datasets and upsampling factors, demonstrating that our method provides the best performance in all the different metrics.

{\bf Run Time:}
Table~\ref{tab:times} compares the average run time for processing a single test image for FRGC $4\times$ upsampling. LSF and YLY are computationally expensive because of the NN search for each of the training images, requiring the run time linear to the number of training images.
MZQ's run time also increases linearly with the number of training images, but it is faster than LSF and YLY since the NN patches are searched only at the specific pixel location. The other algorithms have the run time independent of the number of training images. The algorithms based on feed-forward neural networks, including ours, achieve efficient processing in the test time\footnote{The run time of SRCNN was measured using the Matlab implementation available on the authors' website~\cite{Dong15PAMI}. It would be the fastest with a C++/GPU implementation since their network has fewer arithmetic operations than BCCNN's and ours.}. The run time of our algorithm was measured using Intel i7 CPU (single core implementation) and NVIDIA 780 GTX GPU.

\subsection{Analysis of the Network Architecture} \label{sec:analysis}

\begin{figure}[!tb]
  \centering
  \small
	\begin{tabular}{cccc}	
  \includegraphics[width=.16\textwidth]{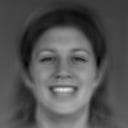}  &
	\includegraphics[width=.16\textwidth]{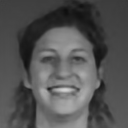} &
	\includegraphics[width=.16\textwidth]{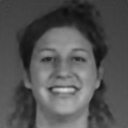} &
	\includegraphics[width=.16\textwidth]{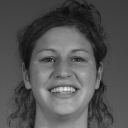} 	\\
  \includegraphics[width=.16\textwidth]{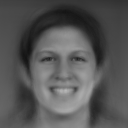}  &
	\includegraphics[width=.16\textwidth]{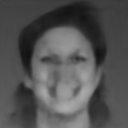}  &
	\includegraphics[width=.16\textwidth]{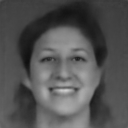}  &
	\includegraphics[width=.16\textwidth]{Figures//gt.png} \\
	(a) GN-Only & (b) LN-Only & (c) GLN & (d) GT \\	
	\end{tabular}
\vspace{-2mm}
  \caption{Qualitative comparison of sub-modules of our network. First and second row show $4 \times$ and $8 \times$ upsampling results respectively. See text for details.}
\label{fig:subnetwork}
\end{figure}

Here we analyze the sub-modules of our network. Figure~\ref{fig:subnetwork} shows $4 \times$ (top row) and $8 \times$ (bottom row) upsampling results using the sub-modules---global upsampling network and local refinement network. These networks are slightly different from the original networks such that (1) we train global detail generation stream (fully connected) of GN to directly produce the high-resolution image, which we call GN-Only; (2) we train LN8 network using only bilinear upsampled low-resolution image as the input, which we call LN-Only. As shown in Figure~\ref{fig:subnetwork}(a), GN-Only produces high quality global details such as symmetry and characteristic facial features, while smoothing out uncommon features (e.g., details on the cheeks and highlights) and producing high frequency artifacts that are not consistent with face patch statistics. Figure~\ref{fig:subnetwork}(b) shows an example of LN-Only result where global details such as characteristic features and symmetry are not preserved (especially for $8 \times$), but the local patch statistics are consistent with face patch statistics (e.g., sharp edges) and local details are preserved. The results of $4 \times$ upsampling is significantly better that $8 \times$ upsampling using LN-Only where resolving patch level ambiguities is easier. The GLN successfully utilizes both global and local cues and produces significantly higher quality results than both sub-modules (Figure~\ref{fig:subnetwork}(c)).

\begin{table}[t]
    \caption{Average run time for FRGC $4\times$ upsampling.}
    \vspace{-4mm}
    \label{tab:times}
    \centering
    \small
        \begin{tabular}[t]{|c|c|c|c|c|c|c|c|}
        \hline
        Method &
        KK     &
        SRCNN &
        LSF        &
        MZQ      &
        YLY   &
        BCCNN &
        GLN   \\
        \hline
        Run Time (seconds)
        & 3.9
        & 1.7
        & 940
        & 1.1
        & 2900
        & \bf 0.005 (GPU)
        & 0.023 (GPU) \\
        & & & & & & \bf / 0.04 (CPU) & / 0.91 (CPU)\\
        \hline
        \end{tabular}
\end{table}

Table~\ref{tab:Variations} shows quantitative comparisons using the sub-modules of our network. These results are consistent with the qualitative comparisons and show that LN-Only produces high quality $4 \times$ upsampling and the results degrade at $8 \times$. We also compare the variants of GLN with different numbers of convolutional layers as shown in Table~\ref{tab:LN}. The results show that the improvement is significant from 4 to 6 convolutional layers but it starts saturating afterwards.

\begin{table}[t]
\centering
\small
\caption{Results using the sub-modules of our network and different configurations (PSNR / SSIM).}
\vspace{-4mm}
\label{tab:Variations}
\begin{tabular}[t]{|c|c|c|}
\hline
Method & FRGC $4\times$  & FRGC $8\times$ \\
\hline\hline
GN-Only                    & 27.36 / 0.801 & 25.78 / 0.752 \\
\hline
LN-Only                     & 29.92 / 0.874 & 25.97 / 0.743 \\
\hline
GLN4                   & 29.59 / 0.867 & 26.35 / 0.765 \\
\hline
GLN6                   & 29.94 / 0.875 & 26.61 / 0.775 \\
\hline
GLN8                      & 30.34 / 0.884 & 26.75 / 0.787 \\

\hline
\end{tabular}
\end{table}

\begin{figure}[t]
\centering
\setlength{\tabcolsep}{0.4mm}
\scriptsize
\begin{tabular}[t]{cccccc}

\includegraphics[width=.15\textwidth]{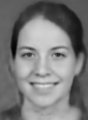} &
\includegraphics[width=.15\textwidth]{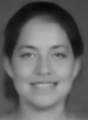} &
\includegraphics[width=.15\textwidth]{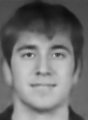} &
\includegraphics[width=.15\textwidth]{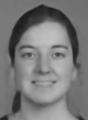} &
\includegraphics[width=.15\textwidth]{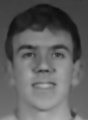} &
\includegraphics[width=.15\textwidth]{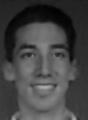} \\

\includegraphics[width=.15\textwidth]{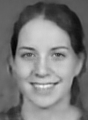} &
\includegraphics[width=.15\textwidth]{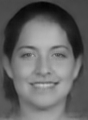} &
\includegraphics[width=.15\textwidth]{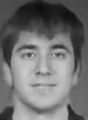} &
\includegraphics[width=.15\textwidth]{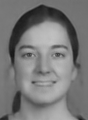} &
\includegraphics[width=.15\textwidth]{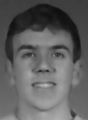} &
\includegraphics[width=.15\textwidth]{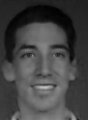} \\

\end{tabular}
\vspace{-2mm}
\caption{Qualitative comparisons of FRGC $8\times$ (left 3 columns) and $4\times$ (right 3 columns) upsampling results using the reconstruction cost only (top row) and after the adversarial fine-tuning (bottom row).}
\label{fig:Adversary}
\end{figure}

\subsection{Adversarial Fine-Tuning}

Figure~\ref{fig:Adversary} compares the results of training the GLN using reconstruction cost only (top row) with fine-tuning the GLN using the combined adversarial loss function (bottom row), as explained in Section~\ref{sec:Training}. The adversarial fine-tuning further improves the visual quality of the generated high-resolution face images where the images are sharper and have more characteristic details. However, this step marginally reduces the PSNR score---$0.01dB$ and $0.25dB$ for $4 \times$ and $8 \times$ respectively. This is expected since the additional adversarial loss does not use the identity of the faces but only evaluates the quality of the generated face images.
%

\begin{figure}[!h]
  \centering
  \small
\begin{tabular}[t]{ccccc}
\includegraphics[width=0.18\textwidth]{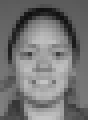}& 
\includegraphics[width=0.18\textwidth]{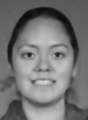}& 
\includegraphics[width=0.18\textwidth]{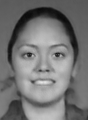}& 
\includegraphics[width=0.18\textwidth]{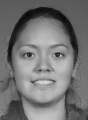}& 
\includegraphics[width=0.18\textwidth]{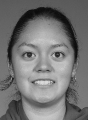}\\ 

\includegraphics[width=0.18\textwidth]{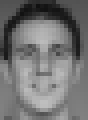}& 
\includegraphics[width=0.18\textwidth]{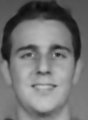}& 
\includegraphics[width=0.18\textwidth]{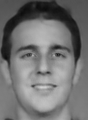}& 
\includegraphics[width=0.18\textwidth]{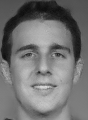}& 
\includegraphics[width=0.18\textwidth]{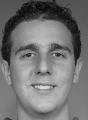}\\ 

\includegraphics[width=0.18\textwidth]{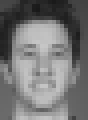}& 
\includegraphics[width=0.18\textwidth]{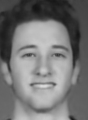}& 
\includegraphics[width=0.18\textwidth]{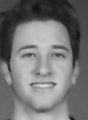}& 
\includegraphics[width=0.18\textwidth]{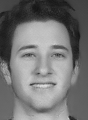}& 
\includegraphics[width=0.18\textwidth]{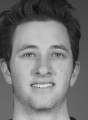}\\ 

\includegraphics[width=0.18\textwidth]{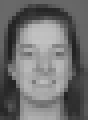}& 
\includegraphics[width=0.18\textwidth]{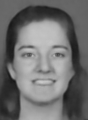}& 
\includegraphics[width=0.18\textwidth]{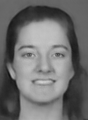}& 
\includegraphics[width=0.18\textwidth]{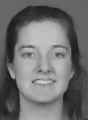}& 
\includegraphics[width=0.18\textwidth]{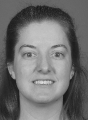}\\ 

\includegraphics[width=0.18\textwidth]{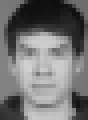}& 
\includegraphics[width=0.18\textwidth]{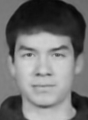}& 
\includegraphics[width=0.18\textwidth]{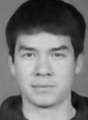}& 
\includegraphics[width=0.18\textwidth]{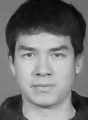}& 
\includegraphics[width=0.18\textwidth]{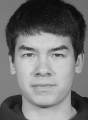}\\

\includegraphics[width=0.18\textwidth]{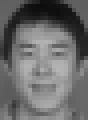}& 
\includegraphics[width=0.18\textwidth]{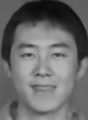}& 
\includegraphics[width=0.18\textwidth]{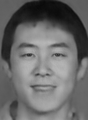}& 
\includegraphics[width=0.18\textwidth]{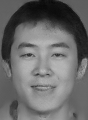}& 
\includegraphics[width=0.18\textwidth]{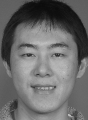}\\ 
Input & $\lambda = 0$ & $\lambda = 10^3$ & $\lambda = 2*10^3$ & Ground Truth \\	
\end{tabular}
\vspace{-2mm}
  \caption{$4 \times$ upsampling results for various weighting factors used in the adversarial loss. }
	 \label{fig:ad4x}
\end{figure}

\begin{figure}[!h]
  \centering
  \small
\begin{tabular}[t]{ccccc}
\includegraphics[width=0.18\textwidth]{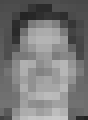}& 
\includegraphics[width=0.18\textwidth]{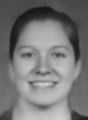}& 
\includegraphics[width=0.18\textwidth]{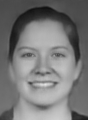}& 
\includegraphics[width=0.18\textwidth]{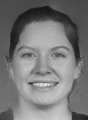}& 
\includegraphics[width=0.18\textwidth]{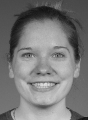}\\ 
\includegraphics[width=0.18\textwidth]{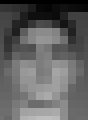}& 
\includegraphics[width=0.18\textwidth]{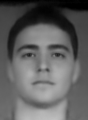}& 
\includegraphics[width=0.18\textwidth]{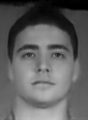}& 
\includegraphics[width=0.18\textwidth]{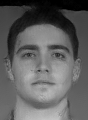}& 
\includegraphics[width=0.18\textwidth]{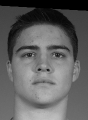}\\ 
\includegraphics[width=0.18\textwidth]{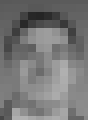}& 
\includegraphics[width=0.18\textwidth]{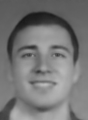}& 
\includegraphics[width=0.18\textwidth]{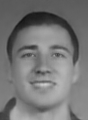}& 
\includegraphics[width=0.18\textwidth]{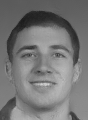}& 
\includegraphics[width=0.18\textwidth]{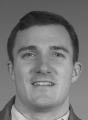}\\ 
\includegraphics[width=0.18\textwidth]{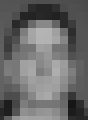}& 
\includegraphics[width=0.18\textwidth]{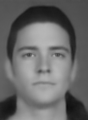}& 
\includegraphics[width=0.18\textwidth]{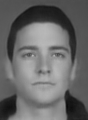}& 
\includegraphics[width=0.18\textwidth]{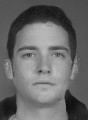}& 
\includegraphics[width=0.18\textwidth]{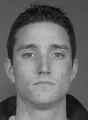}\\ 
\includegraphics[width=0.18\textwidth]{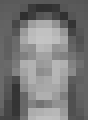}& 
\includegraphics[width=0.18\textwidth]{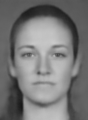}& 
\includegraphics[width=0.18\textwidth]{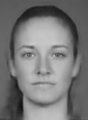}& 
\includegraphics[width=0.18\textwidth]{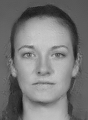}& 
\includegraphics[width=0.18\textwidth]{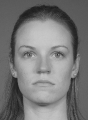}\\ 
\includegraphics[width=0.18\textwidth]{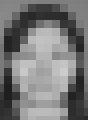}& 
\includegraphics[width=0.18\textwidth]{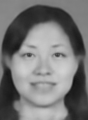}& 
\includegraphics[width=0.18\textwidth]{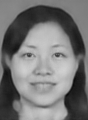}& 
\includegraphics[width=0.18\textwidth]{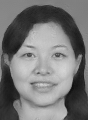}& 
\includegraphics[width=0.18\textwidth]{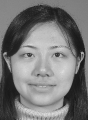}\\ 
Input & $\lambda = 0$ & $\lambda = 4*10^3$ & $\lambda = 8*10^3$ & Ground Truth \\	
\end{tabular} 
\vspace{-2mm}
  \caption{$8 \times$ upsampling results for various weighting factors used in the adversarial loss. }
	\label{fig:ad8x}
\end{figure}

We analyze the effect of the weighting factor, $\lambda$, between the mean-squared loss and the adversarial loss, equation ~\eqref{eq:generator}, on the super-resolution result. The $4 \times$ upsampling results are shown in Figure~\ref{fig:ad4x}, and $8 \times$ upsampling results are shown in Figure~\ref{fig:ad8x}. The weighting factors $\lambda=10^3$  (for $4 \times$ upsampling) and $\lambda=4*10^3$ (for $8 \times$ upsampling) correspond to the results presented in Figure~\ref{fig:Adversary}.  We also present results for $\lambda=2*10^3$ and $\lambda=8*10^3$ for $4 \times$ and  $8 \times$ upsampling respectively. The weighting factor  $\lambda=0$ corresponds to training the network using only the mean-squared loss.

The more we weight the adversarial loss (larger $\lambda$ values), the reconstructed images become sharper and they include more facial details. However, with larger $\lambda$ we also observe some high frequency artifacts.

\subsection{Color Face Upsampling}

\begin{figure}[!h]
  \centering
  \small
\begin{tabular}[t]{ccccc}
\includegraphics[width=0.18\textwidth]{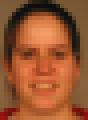}& 
\includegraphics[width=0.18\textwidth]{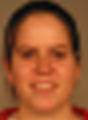}& 
\includegraphics[width=0.18\textwidth]{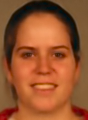}& 
\includegraphics[width=0.18\textwidth]{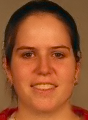}& 
\includegraphics[width=0.18\textwidth]{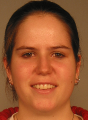}\\ 
\includegraphics[width=0.18\textwidth]{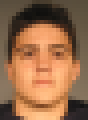}& 
\includegraphics[width=0.18\textwidth]{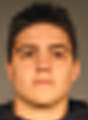}& 
\includegraphics[width=0.18\textwidth]{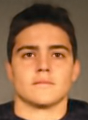}& 
\includegraphics[width=0.18\textwidth]{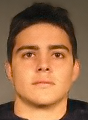}& 
\includegraphics[width=0.18\textwidth]{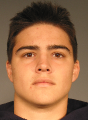}\\ 
\includegraphics[width=0.18\textwidth]{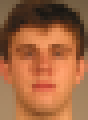}& 
\includegraphics[width=0.18\textwidth]{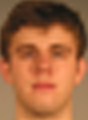}& 
\includegraphics[width=0.18\textwidth]{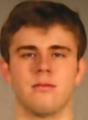}& 
\includegraphics[width=0.18\textwidth]{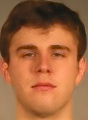}& 
\includegraphics[width=0.18\textwidth]{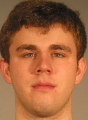}\\ 
\includegraphics[width=0.18\textwidth]{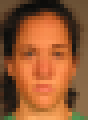}& 
\includegraphics[width=0.18\textwidth]{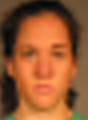}& 
\includegraphics[width=0.18\textwidth]{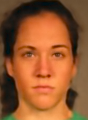}& 
\includegraphics[width=0.18\textwidth]{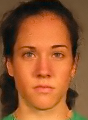}& 
\includegraphics[width=0.18\textwidth]{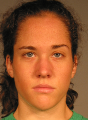}\\ 
\includegraphics[width=0.18\textwidth]{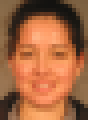}& 
\includegraphics[width=0.18\textwidth]{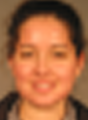}& 
\includegraphics[width=0.18\textwidth]{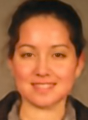}& 
\includegraphics[width=0.18\textwidth]{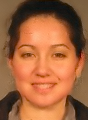}& 
\includegraphics[width=0.18\textwidth]{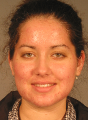}\\ 
\includegraphics[width=0.18\textwidth]{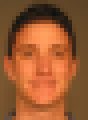}& 
\includegraphics[width=0.18\textwidth]{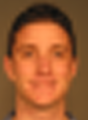}& 
\includegraphics[width=0.18\textwidth]{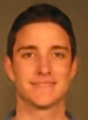}& 
\includegraphics[width=0.18\textwidth]{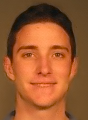}& 
\includegraphics[width=0.18\textwidth]{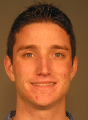}\\ 
Input & Bicubic & GLN & GLN with  & Ground Truth \\	
 &  &  & Adversarial Loss &  \\	
\end{tabular}
\vspace{-2mm}
  \caption{$4 \times$ color upsampling results. }
	 \label{fig:col4x}
\end{figure}

\begin{figure}[!h]
  \centering
  \small
\begin{tabular}[t]{ccccc}
\includegraphics[width=0.18\textwidth]{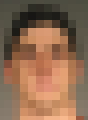}& 
\includegraphics[width=0.18\textwidth]{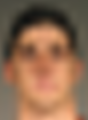}& 
\includegraphics[width=0.18\textwidth]{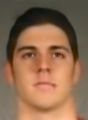}& 
\includegraphics[width=0.18\textwidth]{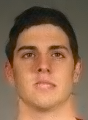}& 
\includegraphics[width=0.18\textwidth]{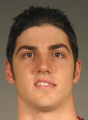}\\ 
\includegraphics[width=0.18\textwidth]{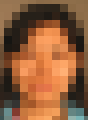}& 
\includegraphics[width=0.18\textwidth]{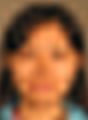}& 
\includegraphics[width=0.18\textwidth]{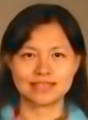}& 
\includegraphics[width=0.18\textwidth]{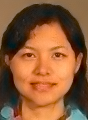}& 
\includegraphics[width=0.18\textwidth]{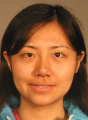}\\ 
\includegraphics[width=0.18\textwidth]{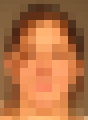}& 
\includegraphics[width=0.18\textwidth]{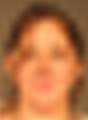}& 
\includegraphics[width=0.18\textwidth]{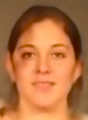}& 
\includegraphics[width=0.18\textwidth]{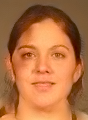}& 
\includegraphics[width=0.18\textwidth]{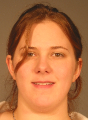}\\ 
\includegraphics[width=0.18\textwidth]{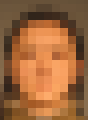}& 
\includegraphics[width=0.18\textwidth]{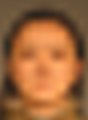}& 
\includegraphics[width=0.18\textwidth]{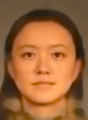}& 
\includegraphics[width=0.18\textwidth]{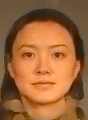}& 
\includegraphics[width=0.18\textwidth]{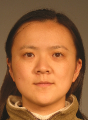}\\ 
\includegraphics[width=0.18\textwidth]{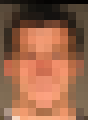}& 
\includegraphics[width=0.18\textwidth]{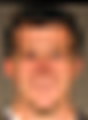}& 
\includegraphics[width=0.18\textwidth]{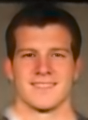}& 
\includegraphics[width=0.18\textwidth]{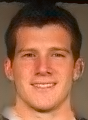}& 
\includegraphics[width=0.18\textwidth]{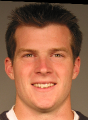}\\ 
\includegraphics[width=0.18\textwidth]{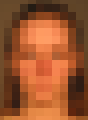}& 
\includegraphics[width=0.18\textwidth]{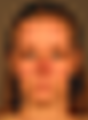}& 
\includegraphics[width=0.18\textwidth]{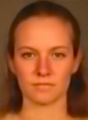}& 
\includegraphics[width=0.18\textwidth]{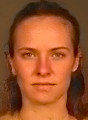}& 
\includegraphics[width=0.18\textwidth]{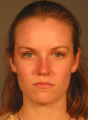}\\ 

Input & Bicubic & GLN & GLN with  & Ground Truth \\	
 &  &  & Adversarial Loss &  \\	
\end{tabular}
\vspace{-2mm}
  \caption{$8 \times$ color upsampling results. }
	 \label{fig:col8x}
\end{figure}

Human vision is not sensitive to chrominance channels (u, v). Therefore, a common procedure for handling color images is to process only the luminance channel (Y) and add bicubic-upsampled chrominance channels to the result. We used the same procedure for obtaining color upsampling results shown in Figures~\ref{fig:col4x} and~\ref{fig:col8x}.

\subsection{Failure Cases}

\begin{figure}[!h]
  \centering
  \small
\begin{tabular}[t]{ccccc}
\includegraphics[width=0.18\textwidth]{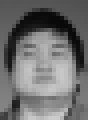}& 
\includegraphics[width=0.18\textwidth]{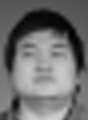}& 
\includegraphics[width=0.18\textwidth]{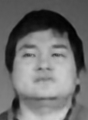}& 
\includegraphics[width=0.18\textwidth]{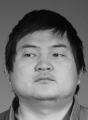}\\ 
\includegraphics[width=0.18\textwidth]{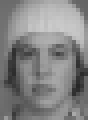}& 
\includegraphics[width=0.18\textwidth]{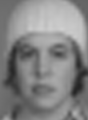}& 
\includegraphics[width=0.18\textwidth]{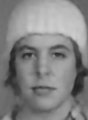}& 
\includegraphics[width=0.18\textwidth]{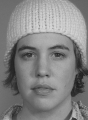}\\ 
\includegraphics[width=0.18\textwidth]{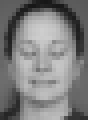}& 
\includegraphics[width=0.18\textwidth]{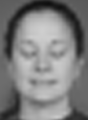}& 
\includegraphics[width=0.18\textwidth]{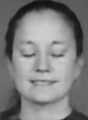}& 
\includegraphics[width=0.18\textwidth]{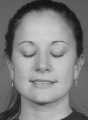}\\ 
\includegraphics[width=0.18\textwidth]{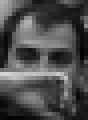}& 
\includegraphics[width=0.18\textwidth]{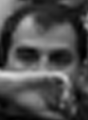}& 
\includegraphics[width=0.18\textwidth]{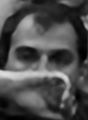}& 
\includegraphics[width=0.18\textwidth]{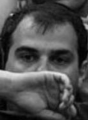}\\ 
\includegraphics[width=0.18\textwidth]{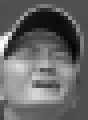}& 
\includegraphics[width=0.18\textwidth]{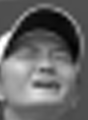}& 
\includegraphics[width=0.18\textwidth]{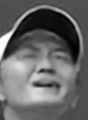}& 
\includegraphics[width=0.18\textwidth]{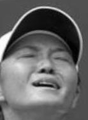}\\ 
\includegraphics[width=0.18\textwidth]{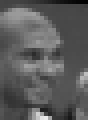}& 
\includegraphics[width=0.18\textwidth]{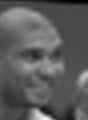}& 
\includegraphics[width=0.18\textwidth]{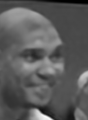}& 
\includegraphics[width=0.18\textwidth]{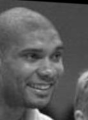}\\ 
Input & Bicubic & GLN & Ground Truth \\	
\end{tabular}
\vspace{-2mm}
  \caption{$4 \times$ upsampling failure examples. }
	 \label{fig:bad4x}
\end{figure}

\begin{figure}[!h]
  \centering
  \small
\begin{tabular}[t]{ccccc}
\includegraphics[width=0.18\textwidth]{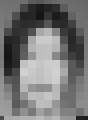}& 
\includegraphics[width=0.18\textwidth]{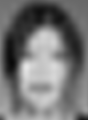}& 
\includegraphics[width=0.18\textwidth]{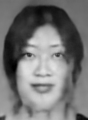}& 
\includegraphics[width=0.18\textwidth]{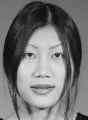}\\ 
\includegraphics[width=0.18\textwidth]{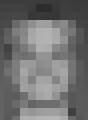}& 
\includegraphics[width=0.18\textwidth]{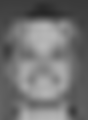}& 
\includegraphics[width=0.18\textwidth]{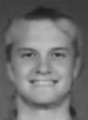}& 
\includegraphics[width=0.18\textwidth]{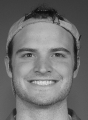}\\ 
\includegraphics[width=0.18\textwidth]{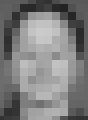}& 
\includegraphics[width=0.18\textwidth]{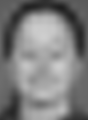}& 
\includegraphics[width=0.18\textwidth]{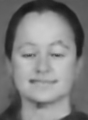}& 
\includegraphics[width=0.18\textwidth]{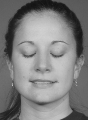}\\ 
\includegraphics[width=0.18\textwidth]{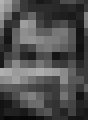}& 
\includegraphics[width=0.18\textwidth]{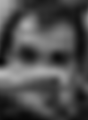}& 
\includegraphics[width=0.18\textwidth]{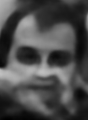}& 
\includegraphics[width=0.18\textwidth]{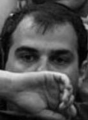}\\ 
\includegraphics[width=0.18\textwidth]{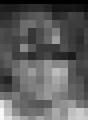}& 
\includegraphics[width=0.18\textwidth]{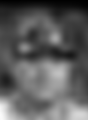}& 
\includegraphics[width=0.18\textwidth]{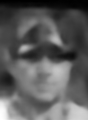}& 
\includegraphics[width=0.18\textwidth]{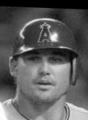}\\ 
\includegraphics[width=0.18\textwidth]{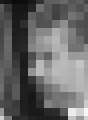}& 
\includegraphics[width=0.18\textwidth]{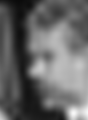}& 
\includegraphics[width=0.18\textwidth]{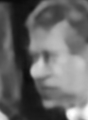}& 
\includegraphics[width=0.18\textwidth]{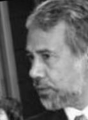}\\ 
Input & Bicubic & GLN & Ground Truth \\	
\end{tabular}
\vspace{-2mm}
  \caption{$8 \times$ upsampling failure examples. }
	 \label{fig:bad8x}
\end{figure}

Our method does not have major failure modes since it does not rely on very precise alignment. The algorithm produces less satisfactory results when there are big variations in pose and facial expression, and/or occlusion. Several less satisfactory results of our algorithm are shown in Figures~\ref{fig:bad4x} and~\ref{fig:bad8x}.

\section{Conclusion} \label{sec:discuss}

We proposed a face hallucination algorithm that produces high quality images even when the input face resolution is very low and the image is captured in an uncontrolled setting. The key element of our algorithm is a deep learning architecture that jointly learns global and local constraints of the high resolution faces. We conducted extensive comparisons with the state-of-the-art algorithms and showed improved performance.

\bibliographystyle{splncs}
\bibliography{reference}
\end{document}